\documentclass[10pt,twocolumn,letterpaper]{article}

\usepackage{cvpr}              % To produce the CAMERA-READY version
\usepackage{graphicx}
\usepackage{amsmath}
\usepackage{amssymb}
\usepackage{booktabs}
\usepackage{cite}

\usepackage[pagebackref,breaklinks,colorlinks]{hyperref}
\usepackage[accsupp]{axessibility}  % Improves PDF readability for those with disabilities.

\usepackage[capitalize]{cleveref}
\crefname{section}{Sec.}{Secs.}
\Crefname{section}{Section}{Sections}
\Crefname{table}{Table}{Tables}
\crefname{table}{Tab.}{Tabs.}

\def\cvprPaperID{2805} % *** Enter the CVPR Paper ID here
\def\confName{CVPR}
\def\confYear{2022}

\begin{document}

\title{A Voxel Graph CNN for Object Classification with Event Cameras}

% \author{Yongjian Deng\\
% Beijing University of Technology\\
% Beijing, China\\
% {\tt\small yjdeng@bjut.edu.cn}
% \and
% Hao Chen\\
% Southeast University\\
% Nanjing, China\\
% {\tt\small haochen593@gmail.com}
% \and
% Hai Liu\\
% Central China Norma University\\
% Wuhan, China\\
% {\tt\small  hailiu0204@ccnu.edu.cn}
% \and
% Youfu Li\\
% City University of Hong Kong\\
% Kowloon, Hong Kong\\
% {\tt\small  meyfli@cityu.edu.hk}
% }

\author{Yongjian Deng$^1$ \qquad Hao Chen$^{2*}$ \qquad Hai Liu$^3$ \qquad  Youfu Li$^{4}$\thanks{: Corresponding author}
\\
\small{$^1$College of Computer Science, Beijing University of Technology}\\
\small{$^2$School of Computer Science and Engineering, Southeast University}\\
\small{$^3$The National Engineering Research Center for E-Learning, Central China Normal University}\\
\small{$^4$Department of Mechanical Engineering, City University of Hong Kong}\\
% {\tt\small yjdeng@bjut.edu.cn, haochen593@gmail.com, hailiu0204@ccnu.edu.cn, meyfli@cityu.edu.hk}
% \and
% Hao Chen\\
% Southeast University\\
% Nanjing, China\\
% {\tt\small haochen593@gmail.com}
% \and
% Hai Liu\\
% Central China Norma University\\
% Wuhan, China\\
% {\tt\small  hailiu0204@ccnu.edu.cn}
% \and
{\tt\small yjdeng@bjut.edu.cn, haochen303@seu.edu.cn, hailiu0204@ccnu.edu.cn, meyfli@cityu.edu.hk}
}
% \footnote{This work was supported by the National Natural Science Foundation of China (Grant No. 61873220, 62102083, 92167102), the Research Grants Council of Hong Kong (Project No. CityU 11213420), and the Science and Technology Development Fund, Macau SAR (File No. 0022/2019/AKP).}
\maketitle
%%%%%%%%% ABSTRACT
\begin{abstract}
\vspace{-1em}
Event cameras attract researchers' attention due to their low power consumption, high dynamic range, and extremely high temporal resolution. Learning models on event-based object classification have recently achieved massive success by accumulating sparse events into dense frames to apply traditional 2D learning methods. 
Yet, these approaches necessitate heavy-weight models and are with high computational complexity due to the redundant information introduced by the sparse-to-dense conversion, limiting the potential of event cameras on real-life applications.
This study aims to address the core problem of balancing accuracy and model complexity for event-based classification models. To this end, we introduce a novel graph representation for event data to exploit their sparsity better and customize a lightweight voxel graph convolutional neural network (\textit{EV-VGCNN}) for event-based classification. Specifically, (1) using voxel-wise vertices rather than previous point-wise inputs to explicitly exploit regional 2D semantics of event streams while keeping the sparsity;
(2) proposing a multi-scale feature relational layer (\textit{MFRL}) to extract spatial and motion cues from each vertex discriminatively concerning its distances to neighbors.
Comprehensive experiments show that our model can advance state-of-the-art classification accuracy with extremely low model complexity (merely 0.84M parameters).\let\thefootnote\relax\footnote{This work was supported by the National Natural Science Foundation of China (61873220, 92167102, 62102083, 62173286, 61875068, 62177018), the Natural Science Foundation of Jiangsu Province (BK20210222), the Research Grants Council of Hong Kong (CityU 11213420), and the Science and Technology Development Fund, Macau SAR (0022/2019/AKP).}

% In specific, (1) we introduce a novel graph representation for event data to exploit their sparsity better and customize a lightweight graph neural network (\textit{EV-VGCNN}) for classification; 
% (2) using voxel-wise vertices rather than traditional point-wise inputs to describe local 2D semantics and incorporate information from more points; (3) proposing a multi-scale feature relational layer (\textit{MFRL}) to extract semantic and motion cues from each vertex discriminatively concerning its distances to neighbors. 

\end{abstract}
\vspace{-2em}
%%%%%%%%% BODY TEXT
\section{Introduction}\label{sec:introduction}

\begin{figure}[]
\centering
\includegraphics[width=0.8\linewidth]{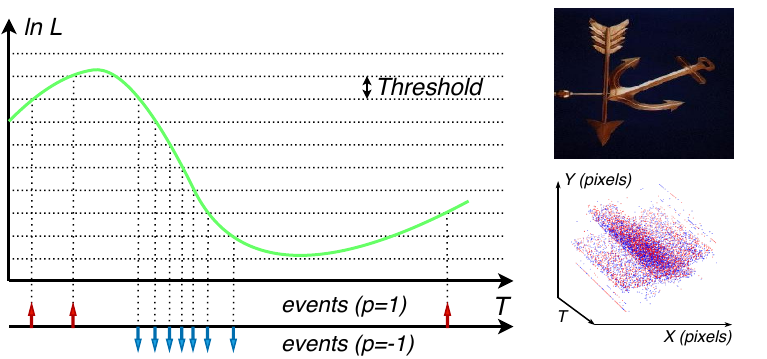}
\caption{\textbf{Left:} A sketch of the working principle of event cameras (the detailed working principle is introduced in supplementary material). Events are produced asynchronously according to the lightness ($\ln{L}$) changes. Red and blue arrows represent positive and negative events, respectively. \textbf{Right:} The RGB image captured from the traditional RGB camera (top) and event signals in the original format produced from an event camera (bottom).}
\label{working_princ}
\vspace{-1.8em}
\end{figure}

% Event cameras, which emulate biological retinas to capture visual changes \cite{lichtsteinerposch, posch2010qvga}, have drawn mounting attention in robotic and computer vision. As illustrated in the left part of Fig. \ref{working_princ}, each pixel output from this sensor (called ``event") is independent and is triggered only when absolute lightness changes in a pixel's location exceed the contrast threshold. As visualized in the right part of Fig. \ref{working_princ}, this novel working mechanism enables their output to be sparse and non-redundant. Consequently, event cameras hold considerable superiority on low power consumption, high response speed, and high dynamic range compared to their frame-based counterparts ($e.g.$, traditional cameras). These advantages allow mobile and wearable devices to perceive and understand real-life scenarios in real-time \cite{gallego2019event}. 

\begin{figure}[]
\centering
\includegraphics[height=0.13\textwidth, width=0.44\textwidth]{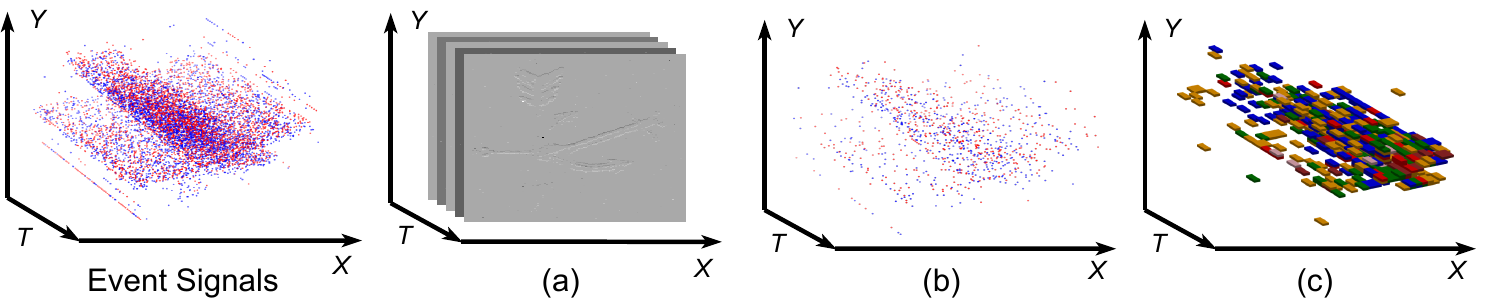}
\caption{Visual comparison of three types of event-based representation: \textbf{(a)} the frame-based representation by integrating events into dense frames; \textbf{(b)} the point-based representation generated by sampling a subset of event signals; \textbf{(c)} representative event voxels selected as vertices in the proposed graph.
}
\label{intro}
\vspace{-1.8em}
\end{figure}

Each pixel of event cameras is independent and only report lightness changes at the correspondent location (Fig. \ref{working_princ}). This novel working principle enables their output to be sparse and non-redundant \cite{lichtsteinerposch, posch2010qvga}. Consequently, event cameras hold advantages of low power consumption, high response speed, and high dynamic range compared to traditional cameras \cite{gallego2019event}. 
% As a fundamental vision task, event-based object classification has always been a emerging research topic. The key problem for event-based object classification is how to tailor a model for the particular format of event data. Motivated by the significant success of deep learning on frame-based vision tasks, developing learning-based algorithms for event data becomes the leading choice. Spiking Neural Networks (SNNs) \cite{maass1997networks} have initially been adopted for processing event data. Although these approaches fit the asynchronous nature of events, the lack of specific hardware supports hinders their potential in real-world applications. Recent works \cite{zihao2018unsupervised, zhu2018ev, gehrig2019end, Cannici_2020_ECCV, 9116961} resort to 2D convolutional neural networks (CNNs) by converting sparse and non-redundant events to dense frames. These representations acquire high performance by taking advantages of the mature 2D CNNs, yet the conversion procedure will introduce noticeable redundant information, $e.g.$, a large number of pixels of dense frames are empty (Fig. \ref{intro} (a)). Besides, these approaches usually need heavy-weight models with large capacities to extract semantic cues from the complex, dense inputs. As a result, they sacrifice the sparsity of event data while limiting the potential of event cameras on mobile or wearable applications.
How to tailor models for event data with the particular format to perform core vision tasks, such as object classification, has been a trending topic. Motivated by the huge success of learning-based methods on vision tasks, developing data-driven approaches for event data becomes a leading choice. For instance, many studies \cite{zihao2018unsupervised, zhu2018ev, gehrig2019end, Cannici_2020_ECCV, 9116961} resort to 2D convolutional neural networks (CNNs) by converting sparse events to dense frames. These works achieve advanced performance utilizing well-pretrained 2D CNNs. Yet, the constructed dense representation and large size models sacrifice the sparsity of event data (Fig. \ref{intro} (a)) while limiting the potential of event cameras on mobile or wearable applications.

To exploit the sparsity of event data and build low-complexity models, recent researchers migrate learning models initially designed for point clouds to event data, such as the works \cite{wang_2018_wacv, Sekikawa_2019_CVPR} migrate models from pointnet-like architecture \cite{qi2017pointnet} and the approaches \cite{mitrokhin2020learning, Graph-based} utilize graph neural networks. 
Although these approaches advance in exploiting the sparsity advantage of event data, a fundamental question has never been investigated: Is point-wise input (taking event points as processing units) proper for event-based vision tasks? % These approaches advance in balancing the performance and model complexity of event-based models. However, a fundamental question has never been investigated, \textit{i.e.}, whether using event points as processing units is proper for event-based vision tasks? 
Intuitively, each point in 3D point clouds can be used as a key point to building the external structure of an object \cite{graph_pc_geometry_coding}. Therefore, these points do well in describing the geometry, which is the key for classifying 3D objects. Instead, event data are more like 2D videos recorded asynchronously. Event-based classification models require the ability to accurately extract 2D semantics from the event data rather than their ``geometry" (Fig. \ref{working_princ}), which usually contains motion cues or motion trajectories. Thus, we believe that using raw events as input is not suitable, as it is difficult for sparse event points to provide decisive features (\textit{e.g.}, local 2D semantics) for event-based models.

To overcome the lacking of characterizing local 2D semantics in the popular point-based solutions, this study proposes a novel voxel-wise representation for event data.
Inspired by the traditional image domain: it is challenging to extract decisive features from images using discrete or discontinuous pixels (like sparse point-wise input). 
Thus, we propose a representation to encode locally coherent 2D semantics by describing the regional events contained in each voxel.
% Our proposed representation is designed to encode locally coherent 2D semantics by describing the regional events contained in each voxel. 
In specific, we build a graph by voxelizing event points, selecting representative voxels as vertices (Fig. \ref{intro} (c)), and connecting them according to their spatio-temporal relationships for further processing.  Each voxel in the graph can be analogous to frame patches of still images that contain essential cues such as local textures and contours \cite{dosovitskiy2020image}, which can help the network recognize 2D scenes effectively. 

% To this end, we propose a novel voxel-wise representation for event data to encode locally coherent semantics by describing the regional events contained in each voxel. In specific, we build a graph by voxelizing event points, selecting representative voxels as vertices as depict in Fig. \ref{intro} (c), and connecting them according to their spatio-temporal relationships. Intuitively, each voxel can be analogous to frame patches of still images that contain essential cues such as local textures and contours \cite{dosovitskiy2020image}, which can help the network recognize 2D scenes effectively. Based on this observation, we believe this novel representation can effectively encode 2D visual features and contribute them to object categorization while keeping the sparsity (Fig. \ref{intro} (c)).
% Moreover, compared to point-wise approaches, since each voxel contains numerous local events, we can scale up semantic and motion representation of the graph to orders of magnitude more event points using a limited number of vertices.

% Besides the newly introduced representation strategy, we propose a lightweight learning architecture (\textit{EV-VGCNN}) with a carefully customized edge embedding method. The proposed model achieves state-of-the-art (SOTA) accuracy while holding surprisingly low model complexity. The critical problem for designing an event-based graph model is how to learn the embeddings for the edges and vertices' features. 
Besides the proposed representation, a lightweight graph-based learning architecture (\textit{EV-VGCNN}) is introduced. 
The critical problem for designing an event-based graph model is how to learn the embeddings for the edges and vertices' features. 
First, we learn a scoring matrix for each vertex according to spatio-temporal geometry and utilize the learned matrix to achieve feature aggregation across its neighbors attentively.
Moreover, for a vertex in the event-based graph, its adjacent neighbors usually convey local spatial messages, while distant neighbors are more likely to carry motion cues or global changes.
Inspired by this variation, we design a multi-scale feature relational layer (\textit{MFRL}) to extract semantic and motion cues from each vertex discriminatively.
Specifically, two learnable blocks in \textit{MFRL} are applied to adjacent and distant neighbors respectively, and the obtained features are aggregated as the joint representation of a vertex. Finally, we cascade multiple \textit{MFRL} modules with graph pooling operations and a classifier as the \textit{EV-VGCNN} to perform end-to-end object classification.  The proposed model achieves SOTA accuracy while holding surprisingly low model complexity. 

The main contributions of this paper are summarized as follows: (1) We introduce a novel method to construct event-based graph representation with correspondence to the properties of event data, which can effectively utilize informative features from voxelized event streams while maintaining the sparse and non-redundant advantage of event data.
(2) We introduce the \textit{MFRL} module composed of several \textit{SFRL}s to discriminatively learn spatial semantics and motion cues from the event-based graph according to spatio-temporal relations between vertices and their neighbors.
(3) Extensive experiments show that our model enjoys noticeable accuracy gains with extremely low model complexity (merely 0.84M parameters).

\section{Related Work}
% Regarding the special format of event data, considerable works customize different handcrafted descriptors from event data for their tasks, such as corner detection \cite{clady2015asynchronous, mueggler2017fast}, edge/line extraction \cite{seifozzakerini2016event, maqueda2018event}, optical flow prediction \cite{ clady2017motion, brosch2015event}, denoising \cite{Probab_denoising} and object classification \cite{lagorce2016hots, sironi2018hats}. These works can be categorized into point-based models, which generally take a single event or a set of regional events as input units for feature extraction and aggregation. Specifically, their model can make task-specific decisions based on single or a few incoming events, thereby taking full advantage of the asynchronous nature of event signals and the temporal information contained therein. Nevertheless, this potential comes with a cost, $i.e.$, the quality of their extracted features is usually sensitive to the noise and scene variation, limiting their generalizability to complex scenarios.

\begin{figure*}[h]
\centering
\includegraphics[height=0.38\textwidth, width=0.9\textwidth]{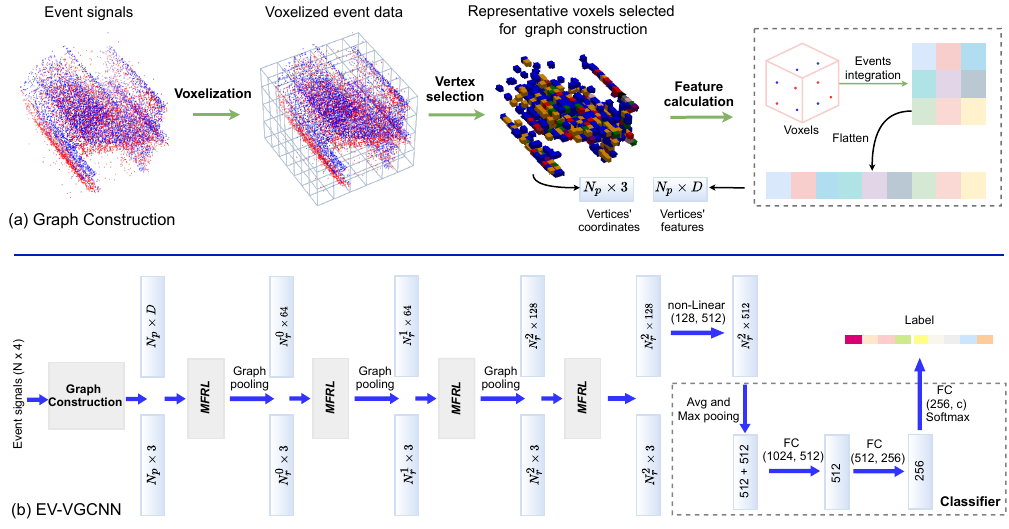}
\caption{(a) \textbf{Graph construction}: We first voxelize event data and then select $N_p$ representative voxels as the graph's input vertices according to the event points number inside each voxel. Finally, we attain features of vertices by integrating their internal events along the time axis. (b) \textbf{EV-VGCNN}: It consists of multiple multi-scale feature relational layers (\textit{MFRL}) followed by graph pooling for learning global features from the input graph, and several linear layers for object classification. The \textit{MFRL} module is to aggregate features for each vertex from k-nearest neighbors. In the figure, non-Linear$(x, y)$ is a block with the input channel of $x$ and output channel of $y$. It contains a linear layer, a Batch Normalization, and a ReLU layer. $N_r^n$ represents the output number of vertices of each graph pooling operation.}
\label{learning architecture}
\vspace{-1.3em}
\end{figure*}

Event-based approaches can be distinguished into two categories: frame-based method and point-based method. 

(i) Frame-based method: integrating event signals into frame-wise 2D representations to directly adopt 2D CNNs for event data. For example, studies such as \cite{maqueda2018event, zhu2018ev, zihao2018unsupervised, 9116961, mvfnet, messikommer2020event} accumulating event points along the time axis to frames \textit{w.r.t} events' polarities, locations and timestamps. To adaptively embed events to frames, \cite{gehrig2019end} and \cite{Cannici_2020_ECCV} introduce learnable neural blocks to weigh the importance of each event in the frame. Besides, The success of methods \cite{ev2vid, Gehrig_2020_CVPR, Hu_2020_Graft, evkdnet, Wang_2021_CVPR} in knowledge transfer also benefits from the frame-based representation. While these frame-based methods achieve encouraging performance on event-based classification, the conversion from sparse points to dense 2D maps introduces much redundant information \cite{mitrokhin2020learning}. As a result, they typically require heavy-weight models to extract high-level features, limiting the potential of event cameras on mobile or wearable applications.

(ii) Point-based method: taking a single event or a set of regional events as input units for feature extraction and aggregation. Early studies falling in this category focus on designing handcraft descriptors for event data to perform multiple vision tasks, such as corner detection \cite{clady2015asynchronous, mueggler2017fast}, edge/line extraction \cite{seifozzakerini2016event, maqueda2018event}, optical flow prediction \cite{ clady2017motion, brosch2015event}, denoising \cite{Probab_denoising} and object classification \cite{lagorce2016hots, sironi2018hats}. However, the quality of their extracted features is usually sensitive to the noise and scene variation, limiting their generalizability to complex scenarios. 
To enable models to respond adaptively to various samples of event data, the development of point-based learning methods is gradually becoming mainstream. There are many visual tasks have been performed successfully, \textit{e.g.}, motion estimation \cite{Sekikawa_2019_CVPR}, motion segmentation \cite{mitrokhin2020learning} and object classification \cite{7010933, amir2017low, diehl2015fast, perez2013mapping, Graph-based, wang_2018_wacv}. Among these approaches, the closest studies to ours are methods inspired by 3D point cloud learning models, such as \cite{wang_2018_wacv, Sekikawa_2019_CVPR, Graph-based, mitrokhin2020learning}, where the RG-CNNs in \cite{Graph-based} acquires superior results over other point-based methods on various datasets. These models are lightweight and show their potential in multiple tasks. However, all these algorithms take the original events as input units for further processing. As stated in Section \ref{sec:introduction}, this input representation is challenging to extract regional 2D semantics effectively from event data. As a result, their models' performance on object classification tasks is far behind the frame-based solutions. Instead of using the original events as input units, this paper proposes a novel voxel-wise representation. Our constructed representation retains more semantic and motion cues while maintaining the sparsity advantage. Moreover, the \textit{MFRL} module allows us to flexibly aggregate features from different neighbors of each vertex in the graph. Consequently, our model outperforms RG-CNNs \cite{Graph-based} with a large margin in terms of accuracy and model complexity. In the supplementary, we also include connections between our models with voxel-based approaches for 3D point cloud learning.

\section{The Proposed Method}
We propose a novel graph construction method for event data and design a lightweight network \textit{EV-VGCNN} for object classification.
The pipeline of our graph-based solution is shown in Fig. \ref{learning architecture}, which can be depicted as follows. $(i)$ The event stream is organized into voxels, where each voxel may contain several events. $(ii)$ We regard each voxel as a vertex in our graph. In specific, the coordinate of a vertex is determined by voxel positions. The feature (temporal and regional semantics) at each vertex is obtained by accumulating event point features inside its corresponding voxel. $(iii)$ We utilize the \textit{EV-VGCNN} for sequentially aggregating each vertex's features to generate global representations. 
In the following, we detail two crucial components in our learning architecture sequentially, including the graph construction (Sec. \ref{Graph construction}) and the \textit{EV-VGCNN} (Sec. \ref{VGCNN}). 

% In the following sections, we first briefly introduce the working principle of event cameras in Sec. \ref{event_signals}. Then, we detail two crucial components in our learning architecture sequentially, including the graph construction (Sec. \ref{Graph construction}) and the \textit{EV-VGCNN} (Sec. \ref{VGCNN}).

% \subsection{Event signals} \label{event_signals}
% As visualized in Fig. \ref{working_princ}, event cameras produce events asynchronously when they detect changes in the log brightness ($\Delta \ln L(x,y,t)$) that exceed the contrast threshold $C$ \cite{gallego2016event} as Eq. \eqref{logBrightness} described.
% \begin{equation}
% \left | \Delta \ln L \right | = \left | \ln L\left ( x, y, t \right ) - \ln L\left ( x, y, t - \Delta t \right ) \right | > C,
% \label{logBrightness}
% \end{equation}
% where $\Delta t$ is the time between the new event and the last event generated at the same location. Each event $e_{i} = (x_{i}, y_{i}, t_{i}, p_{i})$ is triggered at the pixel location of $(x_{i},y_{i})$ at time $t_{i}$ with polarity $p_{i}$ $ \left( p \in \begin{Bmatrix}-1, 1\end{Bmatrix} \right)$. The polarity of an event shows the sign of brightness changes. Precisely, positive events ($p = +1$) represent the lightness increasing ($\Delta \ln L > C$) and negative events ($p = -1$) represent the lightness decreasing  ($\Delta \ln L < - C$).

\subsection{Graph construction} \label{Graph construction}
In this part, we build a directed graph $\mathcal{G} = \begin{Bmatrix}\mathcal{V}, \mathcal{E}\end{Bmatrix}$,  where $\mathcal{V} = \begin{Bmatrix} 1,...,N_{p}\end{Bmatrix}$ and $\mathcal{E}$ represent vertices and edges respectively. Each vertex $\mathcal{V}_i$ has two attributes which are the spatio-temporal coordinate $\mathcal{U}_i \in \mathbb{R}^{1 \times 3}$ and the feature vector $\mathcal{F}_i \in \mathbb{R}^{1 \times D}$. As the network \textit{EV-VGCNN} is capable of finding neighbors and calculate edges' weights for vertices, we only need to determine the coordinate and features of each vertex in the graph construction stage.

\begin{figure*}[h]
\centering
\includegraphics[height=0.26\textwidth, width=0.93\textwidth]{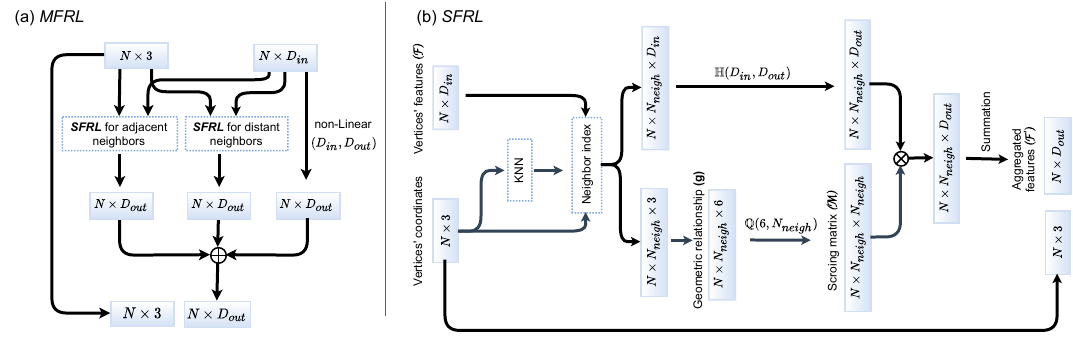}
\caption{Structure of the \textit{MFRL} and its base component \textit{SFRL}. (a) \textbf{\textit{MFRL}}: The \textit{MFRL} is composed of two \textit{SFRL} modules and a shortcut connection, in which the two \text{SFRL} modules are to encode features from adjacent and distant neighbor vertices respectively. This design allows us to explore motion and spatial messages behind event signals flexibly. $\bigoplus$: element-wise addition. (b) \textbf{\textit{SFRL}}: This module realizes the Eq. \eqref{weighting} and Eq. \eqref{aggregation} using neural networks. 
Particularly, the \textit{SFRL} module takes vertices with their coordinates and features as input. For each vertex, the \textit{SFRL} builds edges between the vertex and its $N_{neigh}$ neighbors, computes the scoring matrix for its neighbors then aggregates the features from neighbors using the matrix to obtain the representation of this vertex. $\bigotimes$: matrix multiplication.}
\label{FRL}
\vspace{-1.3em}
\end{figure*}

\noindent{\textbf{Voxelization.}}\indent
Event streams can be expressed as a 4D tuple:
\begin{equation}
\begin{Bmatrix}e_i\end{Bmatrix}_N = \begin{Bmatrix}x_i, y_i, t_i, p_i\end{Bmatrix}_N,
\label{event_stream}
\end{equation}
where $e_i$ is a single event. The first two dimensions $x_i$ and $y_i$ are constrained with the range of the spatial resolution ($\begin{Bmatrix}H, W\end{Bmatrix}$) of event cameras and $t_i$ is the timestamp when the event is triggered. Hence, $(x_i, y_i, t_i)$ represents the spatio-temporal coordinate of an event, and the last dimension $p_i$ can be seen as the attribute. Given an event stream, we can subdivide the 3D x-y-t space into voxels, as shown in Fig. \ref{learning architecture} (a). Considering the value discrepancy between $(x_i, y_i)$ and $(t_i)$, we first normalize the time dimension with a compensation coefficient ${\textit{A}}$ using:
\begin{equation}
\begin{Bmatrix}t_i\end{Bmatrix}_N = \frac{\begin{Bmatrix}t_i - t_0\end{Bmatrix}_N  \times A} {t_{N-1} - t_0}.
\label{Normalize}
\end{equation}
After normalization, event signals encompass 3D space with range $H, W, A$ along the $x, y, t$ axes respectively. Then, we voxelize this 3D space with the size of each voxel as $v_h, v_w$ and $v_a$. The resulting voxels in spatio-temporal space is of size $H_{voxel} = H/v_h$, $W_{voxel} = W/v_w$ and $A_{voxel} = A/v_a$, where each voxel may contain several events. In this work, we refer voxel locations to define the coordinate of each vertex in the graph. Accordingly, the coordinate of \textit{i}-th vertex $\mathcal{U}_i = (x_i^v, y_i^v, t_i^v)$ is with the range of $\begin{Bmatrix}H_{voxel}, W_{voxel}, A_{voxel} \end{Bmatrix}$. For simplicity, we assume that $H, W, A$ are divisible by $v_h, v_w$ and $v_a$.

\noindent{\textbf{Vertex selection.}}\indent
In practice, even though we only consider non-empty voxels as vertices, there are still tens of thousands of vertices that will be contained in a graph. Constructing a graph with all these vertices imposes a substantial computational burden. In addition, due to noise points (also known as hot pixels) produced by event cameras typically occurs isolated without any connections with other events, plenty of vertices may only be composed of single or a few noise points only. Accordingly, taking these uninformative vertices as input would inevitably introduce interference to the whole learning system. Therefore, we propose to keep only representative vertices for graph construction. To this end, we adopt a simple yet effective selection strategy, which is to find $N_p$ vertices with the largest number of event points inside and feed them into the learning system.  On the one hand, this selection strategy can work as a noise filter for input data purification. On the other hand, this procedure can help us save computational costs. Please refer to our supplementary material for more comparisons with different selection approaches.

% \begin{figure*}[h]
% \centering
% \includegraphics[height=0.23\textwidth, width=0.9\textwidth]{Figs/Figure4.pdf}
% % \includegraphics[width=0.9\linewidth]{Figs/Figure4.pdf}
% \caption{Structure of the \textit{MFRL} and its base component \textit{SFRL}. (a) \textbf{\textit{MFRL}}: The \textit{MFRL} is composed of two \textit{SFRL} modules and a shortcut connection, in which the two \text{SFRL} modules are to encode features from adjacent and distant neighbor vertices respectively. This design allows us to explore motion and spatial messages behind event signals flexibly. $\bigoplus$: element-wise addition. (b) \textbf{\textit{SFRL}}: This module realizes the Eq. \eqref{weighting} and Eq. \eqref{aggregation} using neural networks. 
% Particularly, the \textit{SFRL} module takes vertices with their coordinates and features as input. For each vertex, the \textit{SFRL} builds edges between the vertex and its $N_{neigh}$ neighbors, computes the scoring matrix for its neighbors then aggregates the features from neighbors using the matrix to obtain the representation of this vertex. $\bigotimes$: matrix multiplication.}
% \label{FRL}
% \end{figure*}

\noindent{\textbf{Feature calculation for vertices.}}\indent
The performance of graph-based neural networks heavily relies on the quality of each vertex's input features. Since each vertex in our graph contains several event points inside, an appropriate method to encode the features for these points is required.
% Event streams can be seen as a 2D video recorded asynchronously. As an event voxel is generally with a short time span, we argue that integrating its internal event points into a 2D representation along the time axis can well represent 2D appearance semantics for this event voxel. 
Event streams can be seen as a 2D video recorded asynchronously. As an event voxel (vertex) is generally with a short time span, we think it is rational to represent the 2D semantics of voxels simulating the imaging principle of the traditional cameras. That is, accumulating event points into 2D frame patches along the time axis.
Particularly, given a graph with vertices $\begin{Bmatrix}\mathcal{V}_i\end{Bmatrix}_{N_p}$ and coordinates $\begin{Bmatrix}(x_i^v, y_i^v, t_i^v)\end{Bmatrix}_{N_p}$, we can attain 2D features $\mathcal{F}_{i}^{2d} \in \mathbb{R}^{1 \times v_h \times v_w}$ of its $i$-th vertex $\mathcal{V}_i$ as formulated in Eq. \eqref{integrate}.
\begin{equation}
\begin{aligned}
\mathcal{F}_i^{2d} (x, y) = \sum_{i}^{N_v}p_i^{in} \delta(x-x_{i}^{in}, y-y_{i}^{in})t_i^{in},
\end{aligned}
\label{integrate}
\end{equation}
where $N_v$ denotes the number of events inside the vertex $\mathcal{V}_i$. For each internal event, its coordinate ($x_i^{in}, y_i^{in}, t_i^{in}$) is constrained with the size of voxels ($\begin{Bmatrix} v_h, v_w, v_a\end{Bmatrix}$). By linearly integrating the events \textit{w.r.t} their temporal coordinates and polarities, we want to encode the 2D semantics of each vertex while retaining temporal (motion) cues to a certain extent \cite{eventGan, gehrig2019end, zihao2018unsupervised}. Eventually, we flatten the resulting 2D features $\begin{Bmatrix} \mathcal{F}_i^{2d} \end{Bmatrix}_{N_p}$ to obtain feature vectors $\begin{Bmatrix} \mathcal{F}_i \end{Bmatrix}_{N_p} \in \mathbb{R}^{N_p \times D}$ of all vertices in the graph, where $D = v_hv_w$. 

\subsection{EV-VGCNN} \label{VGCNN}
The proposed learning architecture comprises three main components: the multi-scale feature relational layer, the graph pooling operation, and the classifier. In this section, we show how to design them and assemble these modules into the \textit{EV-VGCNN}. 

\begin{figure}[]
\centering
\includegraphics[height=0.14\textwidth, width=0.46\textwidth]{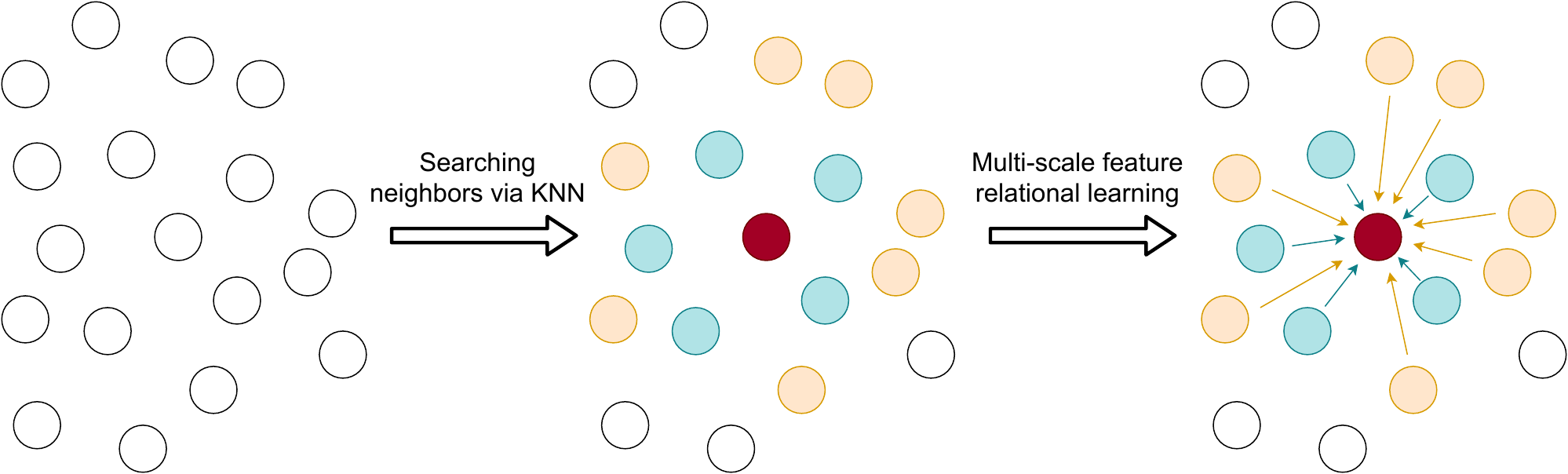}
\caption{An intuitive illustration of how the \textit{MFRL} aggregate features for a vertex from its adjacent and distant neighbors. For a vertex (the red point) in our graph, we firstly determine its $N_{neigh}^{adj}$ (blue points) and $N_{neigh}^{dis}$ (yellow points) neighbors $w.r.t$ distances. Then, we aggregate features from these neighbors utilizing two independent network branches, where yellow arrows and blue arrows represent learned weights from two \textit{SFRL}s.}
\label{intuitiveViz}
\vspace{-1.7em}
\end{figure}

\noindent{\textbf{Multi-scale feature relational layer (\textit{MFRL}).}}\indent
Unlike the traditional 3D point clouds whose coordinates only express geometric messages, event data contains two different types of cues, $i.e.$, 2D spatial messages, and motion information.
For a vertex in the event-based graph, its adjacent neighbors usually carry local spatial cues. In contrast, its distant neighbors are more likely to comprise much motion information. 
Notably, though adjacent and distant neighbors carry motion and spatial cues simultaneously in most cases, the motion and spatial variances between a vertex and its adjacent and distant neighbors are different, $i.e.$, adjacent neighbors hold small and local variance while distant neighbors carry more global changes. Given this disparity, it is difficult to use a shared CNN branch to learn all neighbors. Inspired by the multi-scale learning strategy in \cite{qi2017pointnet++}, we introduce the \textit{MFRL} module to extract motion and semantic messages from vertices distinguishably depending on the distance between vertices and their neighbors. As shown in Fig. \ref{FRL} (a), the \textit{MFRL} consists of one shortcut connection and two single-scale feature relational layers (\textit{SFRL}) to extract correlations from adjacent and distant neighbors respectively. More specifically, for a vertex, we define $N_{neigh}^{adj}$ and $N_{neigh}^{dis}$ as the numbers of its adjacent and distant neighbors, respectively. The results obtained from these three branches are then aggregated as the output. We intuitively illustrate this learning process in Fig. \ref{intuitiveViz}.

We then detail how to design the \textit{SFRL} module.
In contrast to aggregating neighbors' features only using pooling operations like PointNet++ \cite{qi2017pointnet++}, we introduce a scoring matrix computed with correspondence to spatio-temporal relationships between each vertex and its neighbors to achieve feature aggregation attentively. In specific, through the construction procedure depicted in Sec. \ref{Graph construction}, we have obtained features ($\mathcal{F}$) and coordinates ($\mathcal{U}$) of vertices ($\mathcal{V}$) in the graph. The \textit{SFRL} takes these features and coordinates as inputs and is entailed to accomplish three functions: $(i)$ building connections among vertices with edges, $(ii)$ computing the scoring matrix for neighbors of the vertex, and $(iii)$ integrating features from the neighborhood for each vertex. As shown in Fig. \ref{FRL} (b), to achieve these goals, we first utilize the K-Nearest Neighbor algorithm (K-NN) to determine $N_{neigh}$ neighbors for each vertex and link them with edges \cite{point_cloud_inpainting}. The graph includes a self-loop, meaning that each vertex also links itself as a neighbor \cite{KipfW16}. Then, for the \textit{i}-th vertex $\mathcal{V}_i$ with edges $\mathcal{E}_i$ ($\mathcal{E}_i \in \mathbb{R}^{1 \times N_{neigh}}$) and coordinates $\mathcal{U}_i$, the scoring matrix can be calculated using:
\begin{equation}
\begin{aligned}
\mathcal{M}_{i} = \mathbb{Q} (\underset{j:(i,j) \in \mathcal{E}_i}{\mathbb{S}_G} (g_{i, j}) ; W_{m}),
\end{aligned}
\label{weighting}
\end{equation}
where $g_{i, j} = [\mathcal{U}_i, \mathcal{U}_i - \mathcal{U}_j] \in \mathbb{R}^{6}$ represents the geometric relation between a vertex and its neighbors. $\left [ \cdot ,\cdot \right ]$ denotes the concatenation of two vectors.
$\mathbb{S}_G(\cdot)$ is a function that stacks all geometric relations of the vertex's neighbors ($\begin{Bmatrix}g_{i, j} :(i,j) \in \mathcal{E}_i \end{Bmatrix}$) and its output is in $\mathbb{R}^{N_{neigh} \times 6}$. $\mathbb{Q}$ is parameterized by $W_m$ and comprises a linear mapping function, a Batch Normalization, and a Tanh function to explore geometric messages from neighbor vertices. The output $\mathcal{M}_i \in \mathbb{R}^{N_{neigh} \times N_{neigh}}$ is the scoring matrix for $\mathcal{V}_i$, which aims to re-weight features from its neighbors based on spatio-temporal relationships when aggregating the neighbors' features for the central vertex. 
Finally, we formulate the function of aggregating features from neighbors into the vertex as: 
\begin{equation}
\begin{aligned}
\mathcal{F}_{i}^{'} = \sum \mathcal{M}_i ( \mathbb{H} (\underset{j \in \mathcal{E}_i}{\mathbb{S}_F} (\mathcal{F}_j) ; W_{f}) ),
\end{aligned}
\label{aggregation}
\end{equation}
where $\mathbb{S}_F(\cdot)$ is a function that stacks all features ($F_j \in \mathbb{R}^{1 \times D_{in}}$) from neighbor vertices and its output is in $\mathbb{R}^{N_{neigh} \times D_{in}}$. $\mathbb{H}$ is a non-linear transform function with parameters $W_f$ and consists of a linear mapping function, a Batch Normalization and a ReLU. The function $\mathbb{H}$ takes the stacked features as input and produces transformed features in $\mathbb{R}^{N_{neigh} \times D_{out}}$. After that, the scoring map $\mathcal{M}_i$ is utilized to re-weight neighbors' features. We then apply a summation operation on the feature space over all neighbors to generate the aggregated features $\mathcal{F}_{i}^{'} \in \mathbb{R}^{1 \times D{out}}$ for the $i$-th vertex.

\noindent{\textbf{Graph pooling operation.}}\indent
The pooling operation is to reduce the vertex number in the network progressively. The pooling layer in 3D vision models commonly aggregates local messages of each point and then selects a subset of points with the dominant response for the following processing. In our case, the feature aggregation step has been fulfilled by the \textit{MFRL} module. Hence, in our pooling layers, we only need to randomly select vertices from the graph to enlarge the receptive field for the feature aggregation of each vertex. We denote the output number of vertices of the graph pooling operation as $N_r$.

\noindent{\textbf{Classifier.}}\indent
Following the operation used in \cite{qi2017pointnet, qi2017pointnet++}, we apply symmetry functions on the high-level features to achieve a global representation for the input. Specifically, we use max and average pooling operations to process these high-level features respectively, and then concatenate them to form a one-dimensional feature vector. Finally, we feed the global feature vector to three fully connected layers for classification.

\noindent{\textbf{Network architecture.}}\indent
The structure of \textit{EV-VGCNN} is the same for all datasets. As shown in Fig. \ref{learning architecture} (b). We embed four relational learning modules (\textit{MFRL}) into our model to obtain the global representation of an event stream. Besides, we apply the pooling operation after each of the first three \textit{MFRL}s. We further apply a non-linear block consisting of a linear layer, a Batch Normalization, and a ReLU after the fourth \textit{MFRL} and then feed the output feature of this non-linear block to the classifier.

\section{Experimental Evaluation}
In this section, we use several benchmark datasets to evaluate the proposed method on the classification accuracy, the model complexity (measured in the number of trainable parameters), and the number of floating-point operations (FLOPs). Also, please refer to the supplementary material for details about the effectiveness of our method on the action recognition task.

\subsection{Datasets}
We validate our method on five representative event-based classification datasets: N-MNIST (N-M) \cite{orchard2015converting}, N-Caltech101 (N-Cal) \cite{orchard2015converting, fei2006one}, CIFAR10-DVS (CIF10) \cite{li2017cifar10}, N-CARS (N-C) \cite{sironi2018hats} and ASL-DVS (ASL) \cite{bi2019graph}. In general, there are two alternatives to generate these datasets. Particularly, N-M, N-Cal and CIF10 are obtained by recording traditional images displayed on monitors with fixed motion trajectories. This recording method may suffer from artificial noise introduced by the shooting and emulation environments.
On the contrary, N-C and ASL are recorded in the real-world environment using event cameras, which means that the evaluation results on these two datasets should better reflect the performance of event-based models in practice. 
% The statistical comparison of these datasets is shown in Table \ref{datasets}.
We train models separately for each dataset and evaluate the performance of our approach on their testing sets. For those datasets (N-Cal, CIF10, and ASL) without official splitting, we follow the experiment setup adopted in \cite{sironi2018hats, Graph-based}, in which 20\% data are randomly selected for testing, and the remaining is used for training and validation. We average over five repetitions with different random seeds as our reported results.

% \begin{table}[]
% \renewcommand\arraystretch{1.5}
% \caption{Comparison of event-based classification datasets.}
% \begin{center}
% \setlength{\tabcolsep}{2mm}{
% \begin{tabular}{lccc}
% \hline \hline
% \textbf{Dataset} & \multicolumn{1}{l}{\textbf{From images}} & \multicolumn{1}{l}{\textbf{Samples}}  & \multicolumn{1}{l}{\textbf{Classes}}  \\ \hline 
% N-MNIST \cite{orchard2015converting}          & Yes                            & 60000                                                                               & 10                                      \\
% N-Caltech101 \cite{orchard2015converting}    & Yes   & 8246    & 101  \\
% CIFAR10-DVS \cite{li2017cifar10}          & Yes     & 10000   & 10   \\ \hline
% N-CARS \cite{sironi2018hats}           & No   & 24029  & 2    \\
% ASL-DVS \cite{bi2019graph}           & No    & 100800    & 24    \\
% \hline \hline
% \end{tabular}}
% \end{center}
% \label{datasets}
% \end{table}

% compensation coefficient A [v_h, v_w, v_a] input voxel N_p N_{neigh}^{adj} = 10, N_{neigh}^{dis}=15 N_{reduce}=128
\subsection{Implementation details} \label{implementation}

\noindent{\textbf{Graph construction.}}\indent
We fix the compensation coefficient $A$ for all datasets as 9 to normalize event data. We set the input vertex number $N_{p}$ as 512 for N-MNIST and ASL as the objects in them are small size, set $N_p$ as $1024$ for N-CARS, and $N_p = 2048$ for more complex datasets N-Cal and CIF10. 
% According to the spatio-temporal discrepancy across different datasets, we set the voxel size as $(v_h, v_w, v_a) = (2, 2, 1)$ for N-MNIST and $(5, 5, 3)$ for other datasets. 
According to the spatio-temporal discrepancy across different datasets, we set the voxel size as $(v_h, v_w, v_a) = (2, 2, 1)$ for N-MNIST, $(7,7,1)$ for CIF10 and $(5, 5, 3)$ for other datasets. 
Please refer to supplementary materials for details about voxel size settings.

\noindent{\textbf{Network.}}\indent
The values of $N_{neigh}^{adj}$ and $N_{neigh}^{dis}$ for all \textit{MFRL} modules are fixed as 10 and 15 respectively. We set the output vertex number of three graph pooling operations as 896, 768, and 640 for N-Cal, N-C, and CIF10 datasets. For the other two datasets, which only take 512 vertices as input, we remove the pooling operation between \textit{MFRL} modules. We add dropout layers with a probability of 0.5 after the first two fully-connected layers in the classifier to avoid overfitting. Each fully-connected layer is followed by a LeakyReLU and a Batch Normalization except for the prediction layer.

\noindent{\textbf{Training.}}\indent
We train our model from scratch for 250 epochs by optimizing the cross-entropy loss using the SGD \cite{sutskever2013importance} optimizer (except for the ASL) with a learning rate of 0.1 and reducing the learning rate until 1e-6 using cosine annealing. As for the ASL, we experimentally find that using the Adam \cite{kingma2014adam} optimizer with a learning rate of 0.001 and decaying the learning rate by a factor of 0.5 every 20 epochs contributes to better performance. The batch size for training is set to 32 for all datasets. 

\begin{table}[]\small
\renewcommand\arraystretch{0.9}
\begin{center}
\caption{Comparison of the classification accuracy between ours and other point-based methods. $^{\dagger}$ Using our proposed representation as input. Blue and green color indicate the first and second best performance.}
\label{point-based results}
\vspace{-1em}
\setlength{\tabcolsep}{1.8mm}{
\begin{tabular}{lccccc}
\hline 
\textbf{Method}  & \textbf{N-M} & \textbf{N-Cal} & \textbf{N-C} & \textbf{CIF10} & \textbf{ASL} \\ \hline
H-First \cite{7010933}        & 0.712            & 0.054                 & 0.561           & 0.077         & -       \\ 
HOTS \cite{lagorce2016hots}            & 0.808            & 0.21                  & 0.624           & 0.271   & -             \\
% Gabor-SNN \cite{sironi2018hats}      & 0.837            & 0.196                 & 0.789           & 0.245   & -         \\ 
HATS \cite{sironi2018hats}       & 0.991  & 0.642  & 0.902  & 0.524  & - \\ \hline
% G-CNNs \cite{bi2019graph} & 0.985 & 0.630 & 0.902 & 0.515 & 0.875 \\
EventNet \cite{Sekikawa_2019_CVPR} & 0.752 & 0.425  & 0.750 & 0.171 & 0.949 \\
PointNet++ \cite{wang_2018_wacv} & 0.841 & 0.503  & 0.809 & 0.465 & 0.947  \\ 
PointNet++ \cite{wang_2018_wacv}$^{\dagger}$ & 0.955 & 0.621  & 0.907 & 0.533 & 0.956 \\ 
RG-CNNs \cite{Graph-based} & 0.990 & 0.657 & 0.914 & 0.540 & 0.901 \\ \hline
% DGCNN \cite{wang2019dynamic} &  &  &  &  & - \\ \hline
{Ours (w/ SFRL)}     & \textcolor{green}{\textbf{{0.992}}}   & \textcolor{green}{\textbf{{0.737}}}        & \textcolor{green}{\textbf{{0.944}}}  & \textcolor{green}{\textbf{{0.652}}}  & \textcolor{green}{\textbf{{0.962}}}     \\  
\textbf{Ours}     & \textcolor{blue}{\textbf{0.994}}   & \textcolor{blue}{\textbf{0.748}}        & \textcolor{blue}{\textbf{0.953}}  & \textcolor{blue}{\textbf{0.670}}  & \textcolor{blue}{\textbf{0.983}}    \\ \hline 
\end{tabular}}
\end{center}
\vspace{-2.5em}
\end{table}

\subsection{Classification accuracy}

In this section, we report the comparison to two mainstream event-based object classification solutions, namely point-based and frame-based methods, to show the advantages of our model comprehensively. 

\noindent{\textbf{Comparison with point-based methods.}}\indent
As our proposed work also falls under the category of point-based methods, we firstly compare it to SOTA point-based models. As shown in Table \ref{point-based results}, the proposed method outperforms SOTA point-based models consistently. Especially, our approach improves the model's performance by a large margin on four challenging datasets such as N-Cal, N-C, CIF10, and ASL. Surprisingly, when alternative the input from point-wise to voxel-wise representation, the method in \cite{wang_2018_wacv} achieves a significant performance gain, suggesting the effectiveness of our newly introduced event-based representation. Furthermore, Table \ref{model_complexity} illustrates that our method shows huge advantages on the model and computational complexity, $e.g.$, our model can achieve 20 times parameter reduction and are with fewer FLOPs compared to the SOTA method RG-CNNs. We attribute the two-sided improvement to two designs in our model. $(i)$ Our graph is constructed by setting an event voxel instead of a single event point as the vertex. This modification retains more powerful regional semantics than other strategies. The resulting compact and informative inputs largely ease the following network to learn distinguishable features in various scenarios. $(ii)$ The \textit{MFRL} module in our network can exploit semantics and motion cues from each vertex distinguishably concerning its distances to neighbors, allowing us to construct a shallow and lightweight network while achieving SOTA accuracy.  

Moreover, we introduce a baseline model which replaces the \textit{MFRL} with the \textit{SFRL} module in the \textit{EV-VGCNN}. In contrast to learning local and distant cues discriminatively, all neighbors of a vertex are equally treated in the \textit{SFRL} to learn their correlations with shared parameters. For a fair comparison, we set $N_{neigh}$ for each \textit{SFRL} as the summation of $N_{neigh}^{adj}$ and $N_{neigh}^{dis}$ in \textit{MFRL}. As shown, the \textit{MFRL} consistently improves the performance on listed datasets, suggesting that the adopted mutli-scale learning strategy can effectively enhance the discriminativeness of features by considering the spatial-temporal variation across neighbors.

\begin{table}[]\small
\renewcommand\arraystretch{0.9}
\begin{center}
\caption{\small Comparison of different event-based classification models on the model complexity (\#Params) and the number of FLOPs. $^{\dagger}$ GFLOPs = $10^9$ FLOPs. $^{\ddagger}$ Using our proposed representation as input. 
% $^{\S}$ I:frame-based method; II:point-based method. 
$^{\P}$T(CPU) represents that both input generation and inference process run on CPU. T(GPU) means that input construction is on CPU and network inference is on GPU.
}
\vspace{-1em}
\setlength{\tabcolsep}{0.7mm}{
\begin{tabular}{lcccc}
\hline  
\textbf{Method}   & \textbf{\#Params}  & \textbf{GFLOPs}$^{\dagger}$ & \textbf{T(CPU)}$^{\P}$ & \textbf{T(GPU)}$^{\P}$\\ \hline
% Ev-Count \cite{Graph-based}       & 25.61 M    &  3.87    & 21.3 ms & 5.92 ms  \\
EST \cite{gehrig2019end}                & 21.38 M    &  4.28   & 27.1 ms & 6.41 ms     \\
% AMAE \cite{9116961}               & 33.52 M    &  5.46   & 26.4 ms & 6.32 ms      \\
M-LSTM \cite{Cannici_2020_ECCV}    & 21.43 M    &  4.82    &  34.8 ms   & 10.89 ms \\
MVF-Net \cite{mvfnet}             & 33.62 M    & 5.62   & 42.5 ms & 10.09 ms     \\ 
AsyNet \cite{messikommer2020event} & 3.69 M & 0.88 & - & - \\ \hline
% G-CNNs \cite{bi2019graph}       & Point-based         & 18.81M    &  \textbf{0.39}         \\
EventNet \cite{Sekikawa_2019_CVPR}           & 2.81 M    & 0.91   & 9.3 ms & 3.35 ms      \\ %\hline
PointNet++ \cite{wang_2018_wacv}          & 1.76 M    &   4.03  & 174.3 ms & 103.85 ms  \\ %\hline
PointNet++$^{\ddagger}$ \cite{wang_2018_wacv}             & 1.77 M    &  4.17      & 178.4 ms & 107.97 ms\\ %\hline
RG-CNNs \cite{Graph-based}             & 19.46 M    &  0.79    & 1236 ms & -  \\ \hline
\textbf{Ours}                          &  \textbf{0.84 M}    &  \textbf{0.70}  & 26.1 ms & 7.12 ms      \\ \hline  
\end{tabular}}

\label{model_complexity}
\end{center}
\vspace{-2.7em}
\end{table}

\noindent{\textbf{Comparison with frame-based methods.}}\indent
To have a comprehensive analysis, we also compare our method with several representative frame-based approaches as shown in Table \ref{frame-based results}. In particular, MVF-Net has achieved SOTA performance on different classification datasets. From the table, we can see that EST, M-LSTM, and MVF-Net have been consistently improved after using pretrained networks, especially on two datasets (N-Cal, CIF10) converted from traditional images. This is because that the frame-based classification models can take advantage of the weights pretrained on large-scale traditional image datasets ($e.g.$, ImageNet) \cite{8946715}.  However, without utilizing the prior knowledge from conventional images, our approach can still achieve comparable accuracy to these frame-based methods on N-M, N-C, and ASL datasets. More importantly, the proposed method obtains better results on all evaluated datasets than most frame-based methods trained from scratch. These comparisons demonstrate that our architecture and designs are greatly suitable for extracting distinguishable representations from event data. 

\begin{table}[]\small
\renewcommand\arraystretch{0.9}
\begin{center}
\caption{Comparison of the classification accuracy between ours and frame-based methods. $^{\dagger}$ Results are acquired by using the classifier with Resnet-34 \cite{he2016deep} as the backbone. $^{\ddagger}$ We train these approaches from scratch and adopt the same training and testing sets used in this paper. Blue and green color indicate the first and second best performance.}
\vspace{-1em}
\setlength{\tabcolsep}{1.8mm}{
\begin{tabular}{lccccc}
\hline 
\textbf{Method}  & \textbf{N-M} & \textbf{N-Cal} & \textbf{N-C} & \textbf{CIF10} & \textbf{ASL} \\ \hline
\multicolumn{6}{c}{Pretrained on ImageNet \cite{deng2009imagenet}} \\ \hline
% Lifetime \cite{mueggler2015lifetime}  & 0.899               & 0.503               & 0.822          & 0.197  & -   \\
% Synchrony \cite{clady2017motion}  & {0.984}                & {0.619}               & {0.913}           &{0.285}     & -  \\
EST \cite{gehrig2019end}          & 0.991     & 0.837        & 0.925  & 0.749    & 0.991      \\
% AMAE \cite{9116961}              & 0.993   & 0.851  & 0.955  & 0.751    & {0.994}   \\ 
M-LSTM \cite{Cannici_2020_ECCV} $^{\dagger}$ & {0.989}                & {0.857}               & {0.957}           &{0.730}     & 0.992 \\
MVF-Net \cite{mvfnet}        & 0.993 & 0.871 & 0.968 & 0.762 & 0.996 \\ \hline
\multicolumn{6}{c}{Without pretraining} \\ \hline
% Ev-Count \cite{Graph-based} & {0.984}     & {0.637}               & {0.903}           &{0.558}     & 0.886  \\
EST \cite{gehrig2019end} $^{\ddagger}$            & \textcolor{green}{\textbf{0.990}}     & \textcolor{blue}{\textbf{0.753}}           & 0.919  & 0.634    & 0.979      \\
% AMAE \cite{9116961}   $^{\ddagger}$           & 0.983   & 0.694  & 0.936  & 0.620    & \textbf{0.984}   \\
M-LSTM \cite{Cannici_2020_ECCV} $^{\ddagger}$ & {0.986}   & 0.738               & 0.927          & 0.631     & \textcolor{green}{0.980} \\
MVF-Net \cite{mvfnet} $^{\ddagger}$       & 0.981 & 0.687 & 0.927  & 0.599 & 0.971 \\
AsyNet \cite{messikommer2020event} & - & 0.745 & \textcolor{green}{\textbf{0.944}}  & \textcolor{green}{\textbf{0.663}} & - \\ \hline
\textbf{Ours}     & \textcolor{blue}{\textbf{0.994}}   & \textcolor{green}{\textbf{0.748}}      & \textcolor{blue}{\textbf{0.953}}  & \textcolor{blue}{\textbf{0.670}}  & \textcolor{blue}{\textbf{0.983}}     \\ \hline  
\end{tabular}}

\label{frame-based results}
\end{center}
\vspace{-2.7em}
\end{table}

\subsection{Complexity and computation analysis} We follow the calculation method described in \cite{he2015convolutional, howard2017mobilenets, Graph-based, molchanov2019pruning} to compute FLOPs for these methods. Since models' architecture may vary when evaluated on different datasets, we obtain results from these models on the same dataset N-Cal. The model complexity and the number of FLOPs of these frame-based methods are listed in Table \ref{model_complexity}, in which our approach is capable of performing classification with lower computational cost and fewer parameters. Compared to frame-based solutions which introduce much redundant information, our graph network learns decisive features directly from the sparse inputs, thus effectively relieving the learning pressure of the neural network. 
For instance, the 18-channel frame-based representation of samples from the N-Cal dataset with a spatial resolution of $180 \times240$ used in \cite{gehrig2019end} has 777600 input elements, while our proposed graph only contains 2048 ones.

In addition, we compute the averaged computation time for processing each sample in the dataset N-C and list the results in Table \ref{model_complexity}. We implement various methods on a workstation with a CPU (Intel i7), a GPU (GTX 1080Ti), and 64GB of RAM. From the table, we can find that the processing speed of our lightweight model is the same level with frame-based methods ($e.g.$ EST). However, our method shows weakness in computation speed compared to EventNet \cite{Sekikawa_2019_CVPR}, which is developed on PointNet. We attribute this phenomenon to two points. (i) The integration operations for graph construction cost much computation time. (ii) The neighbor searching and feature embedding functions that do not exist in EventNet, though enlarge our model's performance by a large margin, also increase our computation time. Considering the low computational complexity (FLOPs) of our approach, we believe that there will be a large room for our method to speed up with the help of coding optimization. Moreover, though our approach cannot reach the temporal resolution of event data, the processing rate (only need 7.12 ms for each sample, which is equivalent to 140 Hz as frame-rate) is fast enough for most high-speed applications.

\subsection{Ablation study}
In this part, we conduct ablation studies to verify advantages of our voxel-wise graph representations and discuss the impact of hyper-parameters $N_{Neigh}^{adj}$ and $N_{Neigh}^{dis}$ to the system. 
Also, please refer to the supplementary material for details about the robustness of our model to the input vertex density.

\begin{table}[]\small
\renewcommand\arraystretch{0.9}
\begin{center}
\caption{Impact of different value of $N_{neigh}^{adj}$ and $N_{neigh}^{dis}$ on the performance evaluated on the N-Cal dataset.}
\vspace{-1em}
\setlength{\tabcolsep}{1.3mm}{
\begin{tabular}{lccccccc}
\hline 
\multicolumn{1}{l|}{} &\multicolumn{7}{c}{\textbf{Value}} \\ \hline
\multicolumn{1}{l|}{Variants} & A & B & C & D & E & F & G \\ \hline
\multicolumn{1}{l|}{$N_{neigh}^{adj}$}       & 10   & 10 & 10 & 5 & 20 & 5 & 20\\ 
\multicolumn{1}{l|}{$N_{neigh}^{dis}$}     & 15    & 10 & 20 & 15 & 15 & 20 & 5\\ 
\multicolumn{1}{l|}{\textbf{Accuracy}}   & 0.748 & 0.742 & 0.751 & 0.737 & 0.740 & 0.743 & 0.730\\ \hline  %  &  9.54      \\ \hline  \hline
\end{tabular}}

\label{N_neigh_abl}
\end{center}
\vspace{-2.7em}
\end{table}
% \vspace{1cm}

\noindent{\textbf{The value setting for $N_{Neigh}^{adj}$ and $N_{Neigh}^{dis}$.}}\indent 
In this part, we set up a series of experiments on the N-Cal dataset to discuss how the variation of $N_{Neigh}^{adj}$ and $N_{Neigh}^{dis}$ affects the final performance of our model. Results of the controlled experiment are listed in Table \ref{N_neigh_abl}. Comparing the settings $A$, $B$ and $C$, we can find that when $N_{Neigh}^{adj}$ is fixed, a larger value of $N_{Neigh}^{dis}$ results in better performance. Intuitively, when we involve more distant neighbors to aggregate the features of a vertex, a denser neighborhood, carrying more global messages and cross-vertex spatio-temporal relationships, is encoded to aggregate the features for the central vertex. Differently, when we fix the $N_{Neigh}^{dis}$ and change the value of $N_{Neigh}^{adj}$ (e.g., settings $A$, $D$, and $E$) from 10 to 20, the final performance drops considerably. We argue that this is due to only a small number of adjacent neighbors being informative to characterize the local semantic information of a vertex. If a considerable part of adjacent neighbors is actually with large distance, then these ``adjacent" neighbors are difficult to characterize this vertex's local semantics and tend to be interference. For the value chosen of these two hyper-parameters in this work, we firstly fix the summation of neighbors as 25 considering computational budget, then experimentally set $N_{Neigh}^{adj}$ and $N_{Neigh}^{dis}$ as 10 and 15 respectively according to the comparison among settings $A$, $F$ and $G$.

\begin{table}[]\small
\renewcommand\arraystretch{0.9}
\begin{center}
\caption{Comparison between voxel-wise and point-wise graph construction strategies with the same network architecture.}
\vspace{-1em}
\setlength{\tabcolsep}{1.7mm}{
\begin{tabular}{lccc}
\hline 
\textbf{Vertex type}     & \textbf{\#Vertex}  & \textbf{Accuracy} & \textbf{GFLOPs} \\ \hline % & Time (ms) \\ \hline
Original events        & 2048             & 0.565        & 0.63   \\
Original events        & 4096             & 0.601        & 0.66    \\
Original events         & 8192             & 0.619        & 0.72    \\
% \textbf{Event voxels (Ours)}       & 1024       &  0.721       & 0.65 \\
\textbf{Event voxels (Ours)}       & 2048       & 0.748        & 0.70 \\ \hline   %  &  9.54      \\ \hline  \hline
\end{tabular}}

\label{effect_graph}
\end{center}
\vspace{-2.7em}
\end{table}

\noindent{\textbf{Comparison of graph construction strategies.}}\indent
% Our voxel-wise graph construction strategy carries more local cues over the point-wise graph construction method. To validate its advantages, we perform comparisons to a point-based graph construction strategy by selecting a random subset of event points as vertices and assigning the polarity of events to vertices as their features. Our comparison focuses on two concerns: $(i)$ when the number of vertices is the same in two types of graph, whether the proposed voxel-wise graph can provide more distinguishable features for the graph network? $(ii)$ If we increase the number of vertices in the point-wise graph to make its model has the same FLOPs as our voxel-wise graph, will our graph still show significant advantages on the classification accuracy? To answer these questions, we feed the graph inputs generated by these two graph construction methods to the same \textit{EV-VGCNN} architecture and perform comparisons on the N-Cal dataset. 
To validate that our proposed voxel-wise graph is more effective in semantics encoding over point-wise inputs, this section performs comparisons of our voxel-wise graph to point-wise graph \cite{Graph-based}, which is constructed by selecting a random subset of event data as vertices and assigning the polarity of events to vertices as their features. We feed these two graphs to the same model \textit{EV-VGCNN} and test their performance on the N-Cal dataset. The results in Table \ref{effect_graph} show that with the same number of vertices (2048) inside, our graph construction strategy contributes to a significant accuracy gain, indicating that it encodes more informative features from the event data than the point-wise graph. We then increase the vertex number of the point-wise graph to 8192, which is much larger than ours. Even so, our method still has a considerable accuracy leading. We credit these superiorities to that our voxel-wise graph construction strategy enables the vertex to encode local correlations among points, thus carrying a more powerful representation for a point region. In contrast, although the compared point-wise graph reduces the complexity of the input, it leads to a severe loss of local 2D appearance and motion information in each vertex. 

% \begin{figure}[]
% \centering
% \includegraphics[width=0.9\linewidth]{Figs/Figure6.png}
% \caption{Results of the proposed model tested with different number of input vertices in the graph.}
% \label{number_vertices}
% \end{figure}

% \subsubsection{Model robustness to the vertex density}
% Practical applications in real life call for our graph-based classification network to be flexible to different input sizes. It means that one can input our \textit{EV-VGCNN} with an arbitrary number of vertices. This characteristic is significantly helpful to event-based models since real-world scenarios are naturally with different spatial scales or motion trajectories, which entail graphs with distinct numbers of vertices to embed their semantic features correspondingly.
% Therefore, we validate the robustness of our model (trained on graphs with 2048 vertices) to the input vertex density on the N-Caltech101 dataset. Fig. \ref{number_vertices} shows that although the classification accuracy drops dramatically when the number of vertices is fewer than 1536, our approach is capable of keeping a stable performance (performance drops within $1\%$) in a large range ([1536, 2816]) of input vertices, suggesting its robustness to the input vertex density and potential on real-world applications.

%-------------------------------------------------------------------------
\vspace{-0.4em}
\section{Limitation} 
\vspace{-0.4em}
% First, this study is based on the assumption that the motions of objects are continuous. That is, for any time interval, input event data to the model contains events related to objects. When this assumption does not hold, our neighbor searching strategy in Euclidean space may not be able to find proper neighbors for vertices. Second, the potential of our model has not been fully facilitated at the current stage due to the lack of a prior knowledge support from large datasets.

First, this study is based on the assumption that the input event data to the model always relate to objects. When this assumption does not hold, our neighbor searching strategy in Euclidean space may not be able to find proper neighbors for vertices. Second, the potential of our model has not been fully facilitated at the current stage due to the lack of a prior knowledge support from large datasets. Third, the adopted synchronous processing pattern can normally provide more robust global features compared to asynchronous methods \cite{Sekikawa_2019_CVPR, messikommer2020event} while inevitably sacrifice real-time performance.

\vspace{-0.4em}
\section{Conclusion}
\vspace{-0.4em}
In this work, we introduce a novel graph-based learning framework for event data. The proposed voxel-wise graph construction method retains more local information than previous point-wise methods while maintaining the sparsity of event data. Moreover, we tailor a \textit{MFRL} module to explore spatial-temporal relationships between vertices and their neighbors discriminatively. Extensive experiments show the advantages of our designs, as well as the elegant improvement on both accuracy and model complexity achieved by the proposed lightweight \textit{EV-VGCNN}.

% \newpage
%%%%%%%%% REFERENCES
{\small
\bibliographystyle{ieee_fullname}
\bibliography{egbib}
}

\end{document}